\documentclass[10pt,twocolumn,letterpaper]{article}

\usepackage{iccv}
\usepackage{times}
\usepackage{epsfig}
\usepackage{subfigure}
\usepackage{graphicx}
\usepackage{amsmath}
\usepackage{amssymb}
\usepackage{lipsum}
\usepackage{ctable}
\usepackage{multicol, blindtext}
\usepackage{multirow}
\usepackage{acronym}
\usepackage{pifont}
\usepackage[sort]{cite}

\newcommand{\cmark}{\ding{51}}%
\newcommand{\xmark}{\ding{55}}

\definecolor{blue}{rgb}{0.2,0.2,0.7}

\usepackage[accsupp]{axessibility}  

\usepackage[pagebackref=true,breaklinks=true,letterpaper=true,colorlinks,bookmarks=false]{hyperref}

\iccvfinalcopy 

\def\iccvPaperID{6101} 

\begin{document}

\title{How to Design a Three-Stage Architecture for \\ Audio-Visual 
Active Speaker Detection in the Wild}


\author{\parbox{16cm}{\centering
    {\large Okan K\"op\"ukl\"u$^1$, \hspace{0.2cm} Maja Taseska$^2$, \hspace{0.2cm} Gerhard Rigoll$^1$}\\
    {\normalsize
    \vspace{0.17cm}
    $^1$ Technical University of Munich\\
    $^2$ Microsoft Corporation}}
}

\acrodef{STFT}{Short-Time Fourier Transform}
\acrodef{CNN}{Convolutional Neural Network}
\acrodef{AV-ASD}{Audiovisual Active Speaker Detection}
\acrodef{RNN}{Recurrent Neural Network}
\acrodef{DNN}{Deep Neural Network}
\acrodef{MLP}{Multi Layer Perceptron}

\maketitle

\begin{abstract}
Successful active speaker detection requires a three-stage pipeline: (i) audio-visual encoding for all speakers in the clip, (ii) inter-speaker relation modeling between a reference speaker and the background speakers within each frame, and (iii) temporal modeling for the reference speaker. Each stage of this pipeline plays an important role for the final performance of the created architecture. Based on a series of controlled experiments, this work presents several practical guidelines for audio-visual active speaker detection. Correspondingly, we present a new architecture called ASDNet, which achieves a new state-of-the-art on the \mbox{AVA-ActiveSpeaker} dataset with a mAP of 93.5\% outperforming the second best with a large margin of 4.7\%. Our code and pretrained models are publicly available \footnote{https://github.com/okankop/ASDNet}.
\end{abstract}

\section{Introduction}

\begin{figure}[t!]
	\centering
	\includegraphics[width = 0.95\linewidth]{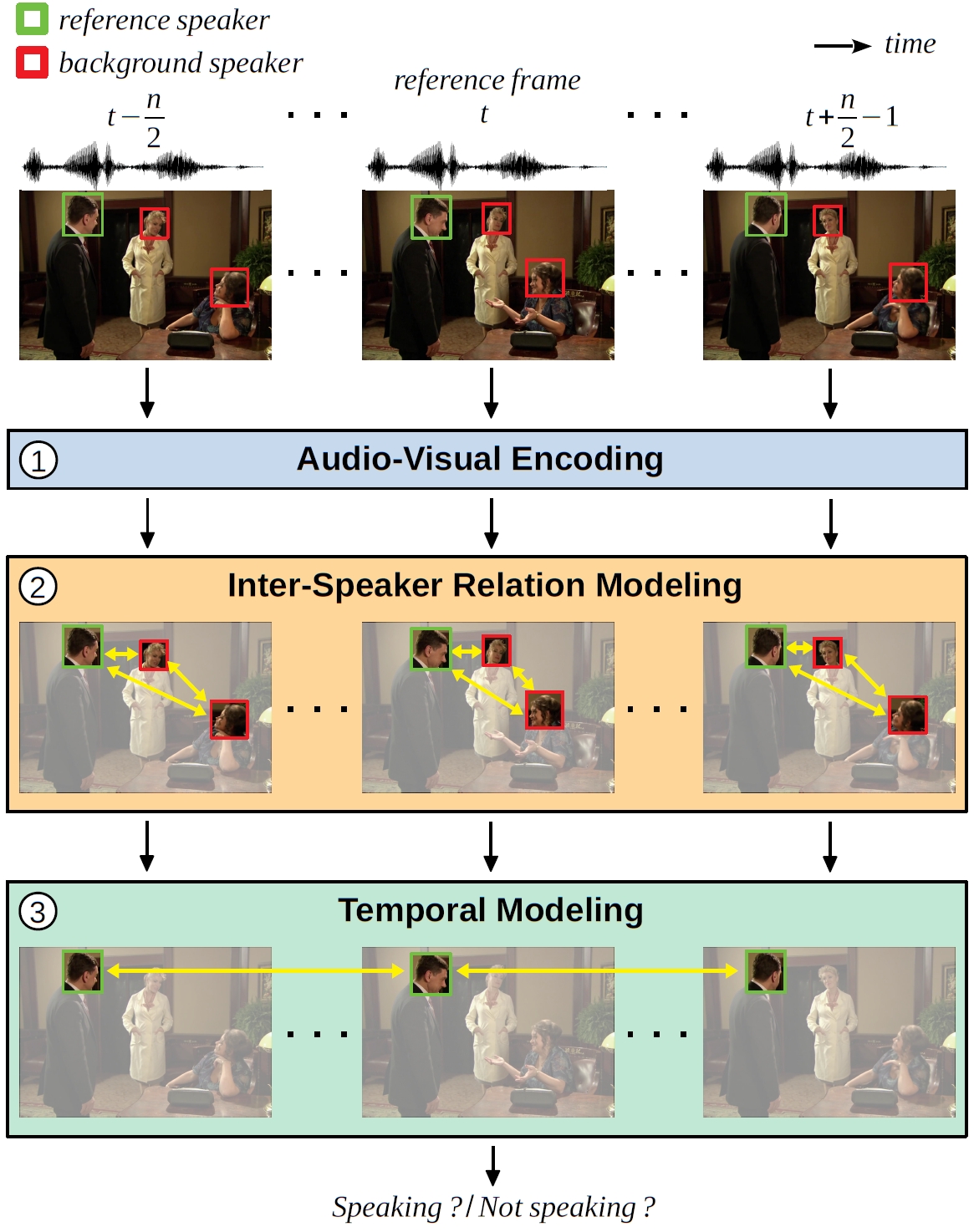}
	\caption{Audio-visual active speaker detection pipeline. The task is to determine if the reference speaker at frame $t$ is \textit{speaking} or \textit{not-speaking}. The pipeline starts with audio-visual encoding of each speaker in the clip. Secondly, inter-speaker relation modeling is applied within each frame. Finally, temporal modeling is used to capture long-term relationships in natural conversations. Examples are from AVA-ActiveSpeaker \mbox{dataset \cite{roth2020ava}.}}
	\label{fig:pipeline}
	\vspace{-0.3cm}
\end{figure}

\label{sec:intro}
Fusion of audio and video modalities has been shown to provide promising solutions to long-standing challenging problems. These include among others,
speaker diarization \cite{gebru2017audio}, biometrics \cite{chung2018voxceleb2}, and action recognition \cite{nagrani2020speech2action, gao2020listen}. Similar to other tasks, \ac{AV-ASD} has also long been studied in literature \cite{cutler2000look, darrell2000audio}. 
A particularly challenging flavor of this problem is \ac{AV-ASD} in the wild, where speech is to be detected and assigned to one of possibly multiple active speakers at each instant in time. Clearly, fusing the complementary discriminative information from audio and video modalities is crucial: visual-only approaches can easily be mistaken by other face/mouth motions such as eating, yawning or emotional expressions. Audio-only approaches, although able to perform source clustering and separation~\cite{hershey2016deep, stoller2018waveunet}, aren't sufficiently robust to count the number of speakers and assign speech to the correct source. This is especially challenging with a single microphone input in acoustically adverse conditions, typically encountered in practice.

Recently, the AVA-ActiveSpeaker dataset \cite{roth2020ava} provided the first large-scale standard benchmark for audio-visual active speaker detection in the wild.  Recent research~\cite{alcazar2020active,leon2021maas} indicates that active speaker detection in the wild requires (i) integration of audio-visual information for each speaker, (ii) contextual information that captures inter-speaker relationships, and (iii) temporal modeling to exploit long term relationships in natural conversation. In this paper, we consolidate this three-stage pipeline for audio visual speaker detection, illustrated in Fig.~\ref{fig:pipeline}, and study the importance of each stage in detail.


\vspace{0.2cm}
\noindent\textbf{Contributions.} We propose a novel three-stage pipeline for audio-visual active speaker detection in the wild. Our architecture, named ASDNet, sets a new state-of-the-art result on AVA-ActiveSpeaker dataset with a 93.5\% mAP, and outperforms the second best method \cite{leon2021maas} with a large margin of 4.7\% mAP (Section \ref{sec:sota}). As part of ASDNet, we propose



\vspace{0.1cm}
\noindent\textbf{(1)} architectures for the audio and video backbones of the audio-visual encoder (Section~\ref{sec:av_enc}), that haven't been previously explored for active speaker detection;

\vspace{0.1cm}
\noindent\textbf{(2)} a simple, yet effective inter-speaker relation modeling mechanism (Section~\ref{sec:isrm});

\vspace{0.1cm}
\noindent\textbf{(3)}
In addition,  we provide detailed ablation study and guidelines for tuning all components of ASDNet. The study includes comparison to the state of the art for the two novel components mentioned above, as well as evaluation of various \ac{RNN} architectures for temporal modeling (Section~\ref{sec:exp_isrm}.). 

\vspace{0.1cm}


\section{Related Work}
\label{sec:relwork}

We present the related work in two parts: (i) audio-visual feature extraction in various applications, and (ii) contributions that address active speaker detection in the wild and its challenges. 

\subsection{Audio-visual feature extraction}

\noindent \textbf{\emph{Audio.}} A common approach to extract features in speech and audio research in different applications, is to use \acp{CNN} and \acp{RNN} with log-Mel or \ac{STFT} spectrograms as inputs~\cite{Ephrat2018}. The popularity of these fixed transforms is due to their success in traditional speech and audio processing and the fact that they extract relevant information from first principles. Furthermore, the image-like configuration of the spectrograms allows employing network architectures well-known from computer vision applications. Particularly, in \ac{AV-ASD}, this allows to use similar audio and video backbone architectures~\cite{leon2021maas,alcazar2020active}.

Based on the interpretation of \acp{CNN} as a data-driven filterbank, researchers have applied \acp{CNN} directly on the audio waveforms to capture discriminative information for the task at hand \cite{dieleman2014end, lee2017raw}. Such an approach in the context of \ac{AV-ASD} has been used for an audio backbone in~\cite{Ariav2019}. However, these approaches need much more data and computational resources that the ones exploiting spectrograms.
With the goal to exploit the best from both worlds, researchers have come up with learnable, but yet constrained transformations of raw audio data. Examples include Harmonic \acp{CNN} used for music tagging, and the SincNet architecture proposed in~\cite{ravanelli2018speaker}. The latter was successfully used in several audio applications ~\cite{kurzinger2020lightweight,mittermaier2020small,liu2020multichannel}. To our best knowledge, this promising architecture hasn't been used in the context of \ac{AV-ASD}. 

\vspace{.2cm}
\noindent {\emph{\textbf{Video.}}} Active speaker detection using only video modality can be viewed as action recognition task. Prior to \acp{CNN}, action recognition research was dominated by hand-crafted features~\cite{laptev2005space, laptev2008learning, wang2013action}, combined with Fisher Vector representations~\cite{perronnin2010improving} or Bag-of-Features histograms~\cite{csurka2004visual}. Ever since AlexNet~\cite{krizhevsky2012imagenet} won the ImageNet Challenge~\cite{russakovsky2015imagenet}, hand-crafted features were mostly abandoned in favor of features extracted by CNNs. This trend extended to video analysis tasks as well, including action recognition. Initially, due to the absence of a large-scale video dataset, architectures for action recognition could benefit from pretraining on the very-large ImageNet dataset~\cite{deng2009imagenet}. 
The first intuitive approach was to treat video frames as multi channel input to 2D-CNNs \cite{simonyan2014two, karpathy2014large}. Other approaches include extraction of frame-level features with a 2D-CNN, followed by a spatiotemporal modeling mechanism \cite{kopuklu2019comparative}.

With the availability of large-scale video datasets such as Kinetics \cite{carreira2017quo}, Moments-in-Time \cite{monfort2019moments}, and Jester \cite{materzynska2019jester}, 2D-CNNs were replaced by 3D-CNNs, to better capture temporal information and motion patterns within video frames. A 3D-CNN architecture was first proposed in~\cite{ji20123d} by Ji \etal. Since then, many 3D-CNN architectures for video recognition tasks followed, such as C3D~\cite{tran2015learning}, I3D~\cite{carreira2017quo}, P3D~\cite{qiu2017learning}, R(2+1)D~\cite{tran2018closer}, SlowFast~\cite{feichtenhofer2019slowfast}, etc. In~\cite{hara2018can}, the effect of dataset size on performance is investigated for several 3D-CNN architectures. Inflated versions of popular resource-efficient 2D-CNN architectures are analyzed for video classification tasks in \cite{kopuklu2019resource}. In this work, we explore variants of 3D-CNNs for the \ac{AV-ASD} task.

\vspace{.2cm}
\noindent \textbf{\emph{Fusion.}} The extracted modality-specific features can be combined at data level~\cite{kopuklu2018motion}, feature level~\cite{miao2017multimodal} or decision level~\cite{simonyan2014two}. The fusion that we apply in this work can be considered as feature level fusion, since we keep processing fused features at inter-speaker relation modeling and temporal modeling mechanisms afterwards.

\subsection{Active speaker detection in the wild}
\label{sec:rel_asd}
Audio-visual active speaker detection is a specific case of source separation \cite{wang2019voicefilter, chakravarty2015s}, where audio and visual signals are leveraged jointly to assign a speech segment to its speaker. For this task, initial approaches \cite{cutler2000look, darrell2000audio} use datasets collected in controlled environments. With the availability of AVA-ActiveSpeaker dataset \cite{roth2020ava}, the research community was able to shift towards active speaker detection in the wild. 

Audio-visual feature extraction is the first step in top-performing frameworks for active speaker detection \cite{roth2020ava, alcazar2020active, leon2021maas, chung2019naver, zhangmulti}. A two-backbone approach has established itself as a standard architecture due to its versatility~\cite{simonyan2014two}. With a good audio-visual feature extraction and \ac{RNN}-based temporal modeling, the authors in~\cite{chung2019naver} achieved competitive performance on the AVA-ActiveSpeaker dataset. Temporal modeling constitutes an integral part of recent active speaker detection pipelines \cite{roth2020ava, alcazar2020active, chung2019naver, zhangmulti}. Often neglected is the context information that can be obtained by modeling inter-speaker relationships. Researchers have only recently proposed methods to exploit the context information\cite{alcazar2020active,leon2021maas}.


\section{Methodology}
\label{sec:method}

Drawing inspiration from the insights in recent research, we seek to establish a general pipeline that incorporates audio-visual encoding, inter-speaker (context) modeling, and temporal modeling. By designing an appropriate architecture for each component, we are able to exceed the state-of-the art performance on the AVA-ActiveSpeaker dataset. 

\begin{figure}[t!]
	\centering
	\includegraphics[width = 1.0\linewidth]{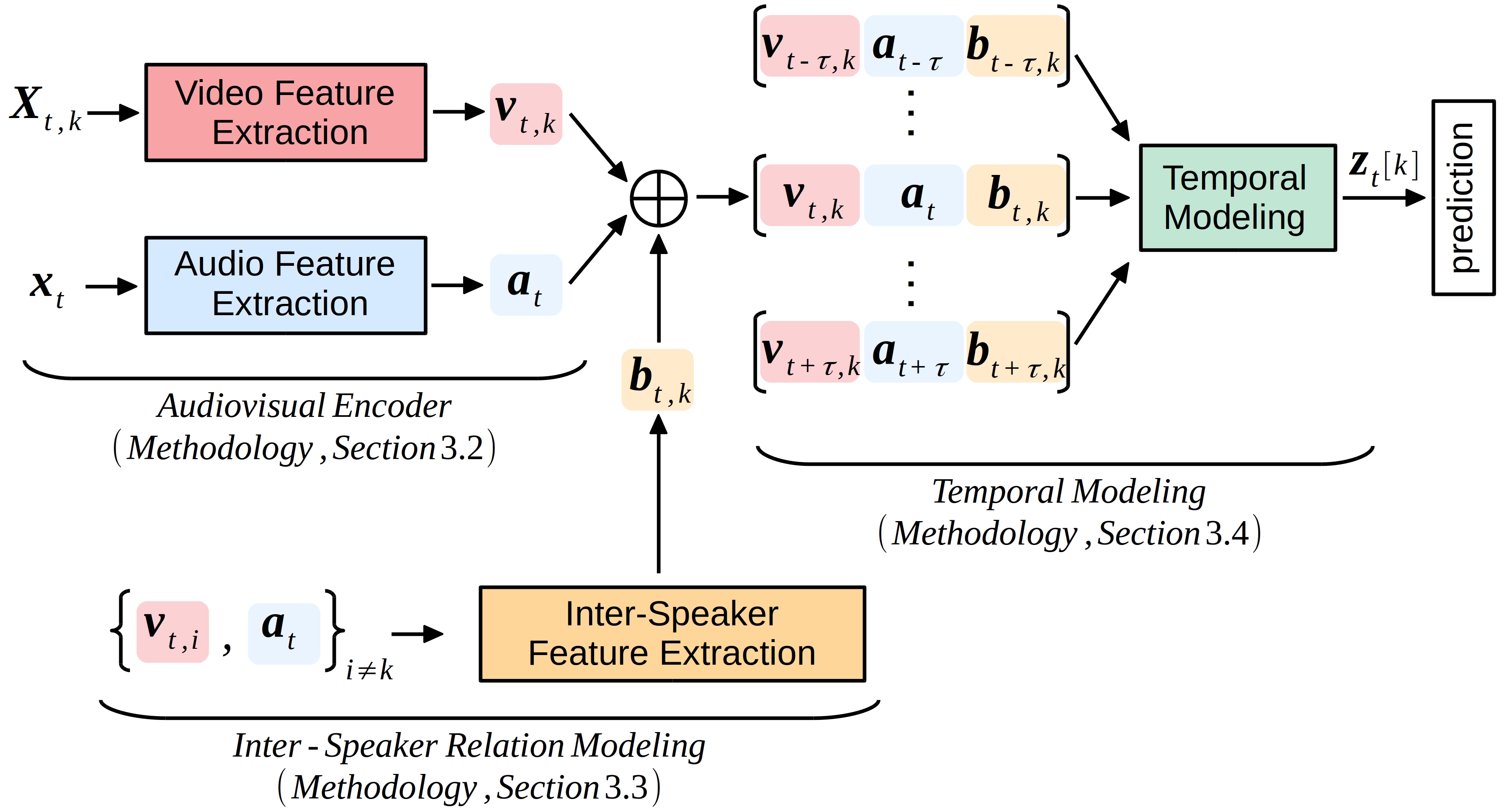}
	\caption{Overview of the three-stage pipeline in ASDNet.} 
	\label{fig:methodology_structure}
\end{figure}

\subsection{Notation and Overview}
\label{sec:overview}

Let $K$ denote the total number of speakers in a given clip. The data available to the active speaker detection system at time $t$ is a set $ \mathcal{X}_t = \lbrace \boldsymbol{X}_{t,1}, \boldsymbol{X}_{t,2}, \ldots \boldsymbol{X}_{t,K}, \boldsymbol{x}_t \rbrace$, where $\boldsymbol{X}_{t,k} \in \mathbb{R}^{n \times 3 \times d_h \times d_w}$ is a tensor of face crops corresponding to the $k$-th speaker. The height and width of the face crops are denoted by $d_h$ and $d_w$, 3 is the RGB channels and $n$ is the number of consecutive face crops centering time instant $t$. The vector $\boldsymbol{x}_t$ contains the samples of the audio track corresponding to the duration of the video input. Given the input data, the objective is to produce a binary vector $\boldsymbol{z}_t$, where $\boldsymbol{z}_t[k] = 1$ if the $k$-th speaker is detected as \emph{speaking} at time frame $t$, and $\boldsymbol{z}_t[k] = 0$ otherwise.

A high-level overview of our pipeline that maps the raw data $\mathcal{X}_t$ to the predictions $\boldsymbol{z}_t$ is illustrated in Fig.~\ref{fig:methodology_structure}. Next, in Sec.~\ref{sec:av_enc}-\ref{sec:temp_mod}, we zoom in on the design of the three pipeline components. In Sec.~\ref{sec:training}, we discuss the training strategy that enables an end-to-end inference: from face crops and an audio waveform, to a prediction \emph{speaking} or \emph{not speaking} for each speaker in the video clip.

\subsection{Audio-Visual Encoder Architecture}
\label{sec:av_enc}

\begin{figure}[t!]
	\centering
	\includegraphics[width = 0.93\linewidth]{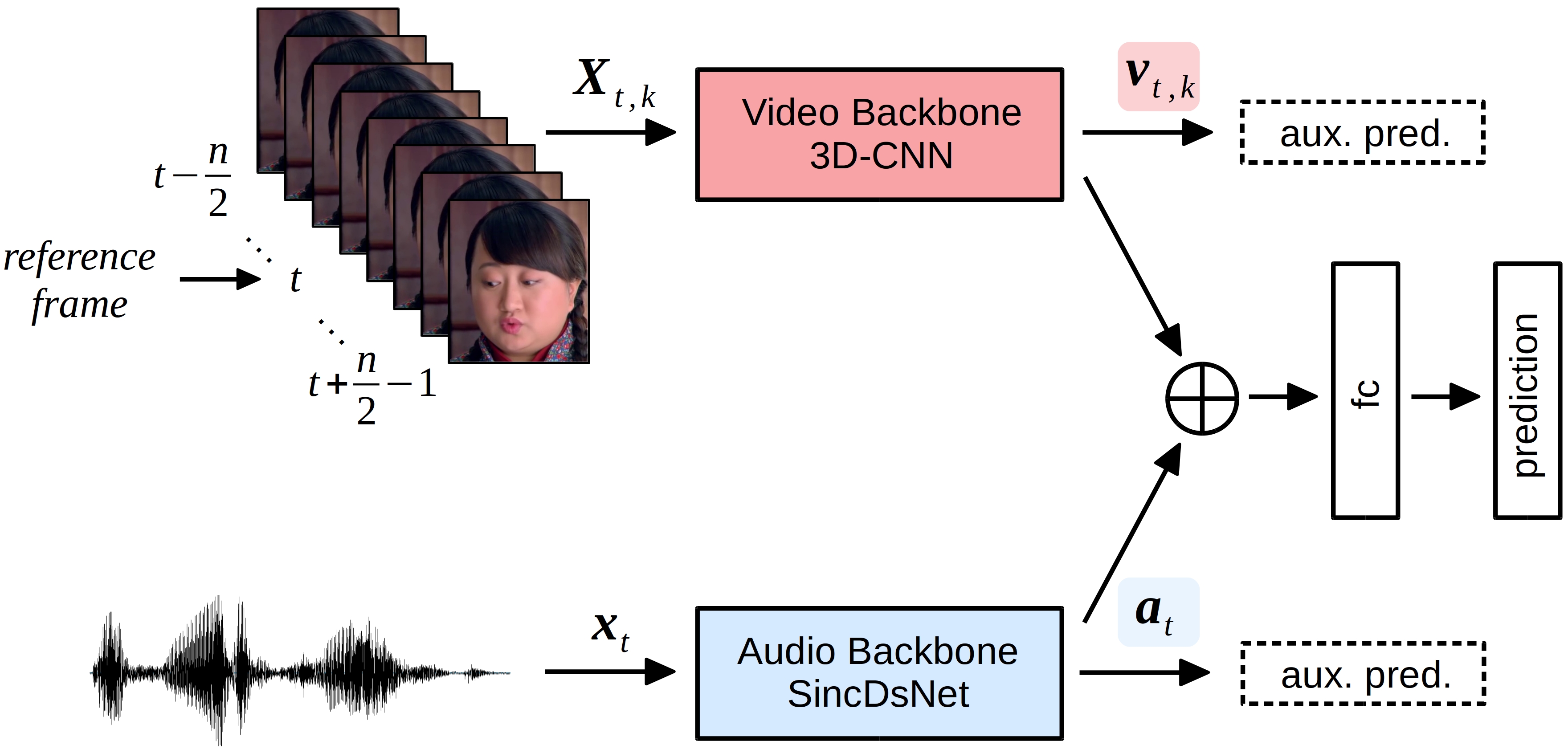}
	\caption{Audio-visual encoder architecture. Visual input $\boldsymbol{X}_{t,k}$ and audio input $\boldsymbol{x}_{t}$ are fed to the respective backbones to produce features $\mathbf{v}_{t,k}$ and $\mathbf{a}_{t}$. A concatenated feature vector $\mathbf{v}_{t,k} \bigoplus \mathbf{a}_{t}$ is fed to a fully connected layer which produces a prediction if speaker $k$ is speaking at time $t$. Prediction heads are removed after training and are not part of the global picture in Fig.~\ref{fig:methodology_structure}.} 
	\label{fig:encoding}
\end{figure}

Our audio-visual encoder is illustrated in Fig.~\ref{fig:encoding}. The stack of face thumbnails $\boldsymbol{X}_{t,k}$ consists of $n$ frames, $X_{t- \frac{n}{2} ,k}, \ldots, X_{t,k}, \ldots,  X_{t+\frac{n}{2}-1,k}$, and the size of the audio input vector $\boldsymbol{x}_t$ is determined by the number of video frames, the video frame rate, and the audio signal sampling rate. The encoder produces an embedding vector by concatenating the modality-specific embeddings
\begin{equation}
     \mathbf{v}_{t,k} =f_v(\boldsymbol{X}_{t,k}; w_v), \hspace{.2cm} \mathbf{a}_{t} =f_a(\boldsymbol{x}_t; w_a),
\end{equation}
where $f_v$ and $f_a$ are neural networks with trainable parameters $w_v$ and $w_a$, respectively.

The concatenated features $\mathbf{v}_{t,k} \bigoplus \mathbf{a}_{t}$ are fed into a fully connected layer to get final predictions. To train the audio-visual encoder, we apply cross-entropy loss between the predictions and ground-truth labels. To ensure that consistent discriminative features are extracted from both modalities, we apply auxiliary classification networks after each backbone, following previous works~\cite{roth2020ava, alcazar2020active, leon2021maas}. 
The auxiliary networks are also trained with cross-entropy loss. The final loss becomes $L_{final} = L_{av} + L_a + L_v$. After training is completed, supervision heads are discarded and only the audio-visual backbone is used to extract features $\mathbf{v}_{t,k}$ and $\mathbf{a}_{t}$ for all speakers and time instants.

While the described high-level architecture is similar to that of existing audio-visual encoders~\cite{roth2020ava, alcazar2020active, leon2021maas}, our contribution lies in the choice and design of the video and audio backbones, discussed next.
\vspace{.2cm}

\textbf{Video backbone.} Movements of mouth and facial muscles are indicative of active speaking. Hence, to fully exploit the available video data, it is important to accurately model motion patterns. To this end, we propose using a 3D-\ac{CNN} as the visual encoder function $f_v$, in contrast to the state-of-the-art approacches that apply 2D-\acp{CNN}\cite{leon2021maas, alcazar2020active, roth2020ava, chung2019naver, zhangmulti}. 
As part of our study, we experimented with various resource-efficient and high-performance 3D-\ac{CNN} architectures \cite{kopuklu2019resource} and found 3D-ResNeXt-101 to be the best performing candidate for our video backbone. Further insights from our investigation are discussed in Section~\ref{sec:exp-ave}.  

\begin{figure}[t!]
	\centering
	\includegraphics[width = 0.75\linewidth]{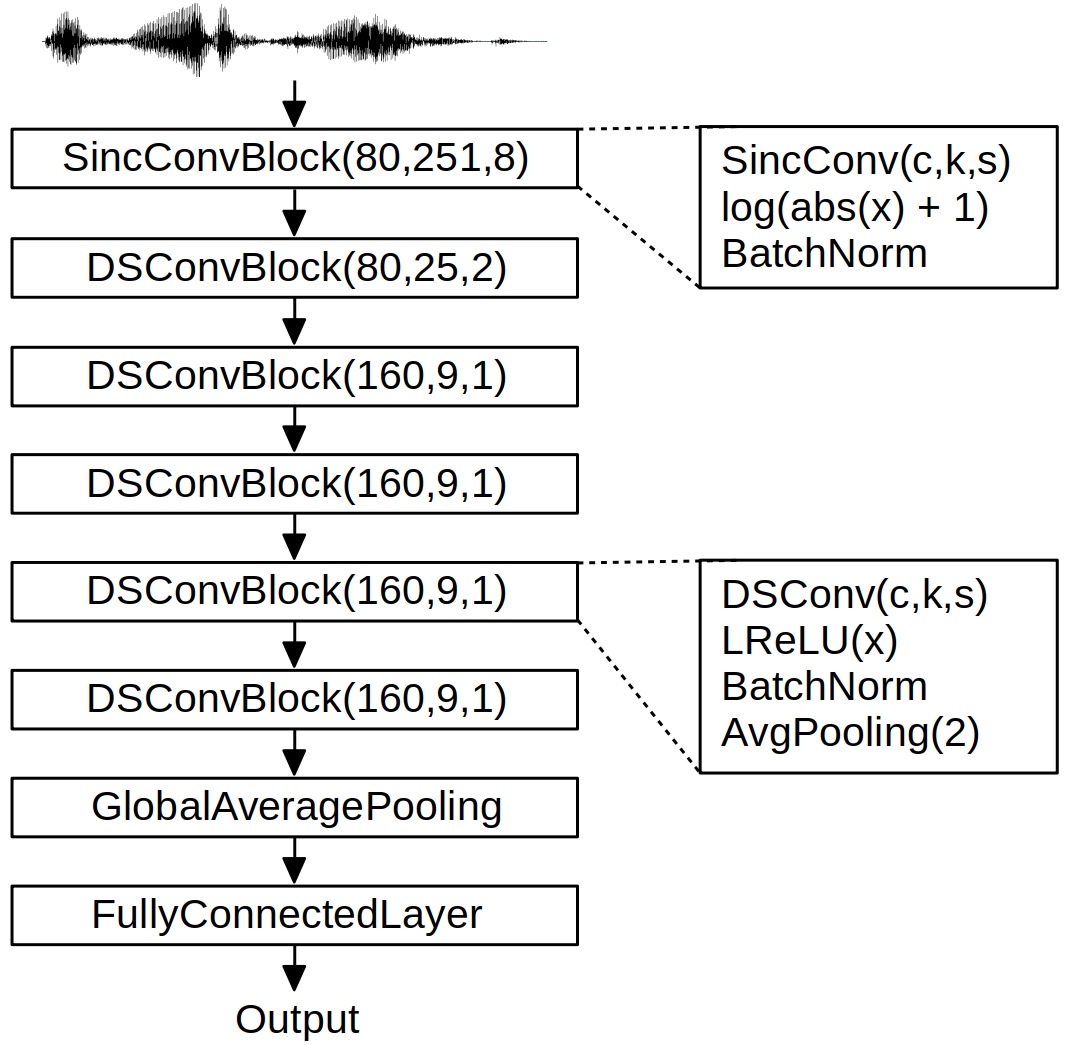}
	\caption{Audio encoder utilizing Sinc Convolutions (SincConv) and Depthwise Separable Convolutions (DSConv). The convolution parameters, \textit{c, k, s} corrspond to the number of output channels, kernel size, and stride, respectively.}
	\label{fig:sincdsnet}
\end{figure}

\vspace{.2cm}
\textbf{Audio backbone.} For the audio encoding backbone, the majority of existing \ac{AV-ASD} approaches~\cite{roth2020ava, alcazar2020active, leon2021maas, chung2019naver, zhangmulti} extract Mel Frequency Cepstral Coefficients (MFCC) from the raw signal, and use the MFCCs as input to 2D-CNNs. In contrast, we propose using an audio backbone architecture that directly operates on raw audio signal via sinc convolutions \cite{ravanelli2018speaker}. In this manner, the system doesn't require a dedicated filterbank and directly exploits all available audio information. This is not the case in existing approaches, where phase information is often discarded after the filterbanks. After sinc convolutions, we apply log-compression, i.e., \mbox{$y = log(abs(x) + 1)$}. This non-linearity has been effective in other raw audio processing tasks as well~\cite{kurzinger2020lightweight, zeghidour2018learning}. The features extracted by the sinc-convolutions are used as input to Depthwise Separable Convolutional (DSConv) blocks with \mbox{Leaky-ReLU} nonlinearity~\cite{xu2015empirical}. Our full audio encoder architecture, referred to as \emph{SincDSNet}, is shown in Fig.~\ref{fig:sincdsnet}. Features after the global average pooling are extracted as the audio \mbox{features $\mathbf{a}_{t}$}. The advantage of the proposed raw-audio backbone over existing feature-based backbones is experimentally demonstrated in Section~\ref{sec:exp-ave}.

\subsection{Inter-Speaker Relation Modeling (ISRM)}
\label{sec:isrm}

The audio-visual encoder extracts features for each individual speaker separately - the features for speaker $k$ do not contain visual information from the remaining speakers in the frame. However, features belonging to background speakers contain complementary information that improves the system performance, as shown in~\cite{alcazar2020active}. In this paper, we propose a method to aggregate information from the background speakers efficiently.

\begin{figure}[t!]
	\centering
	\includegraphics[width = 0.85\linewidth]{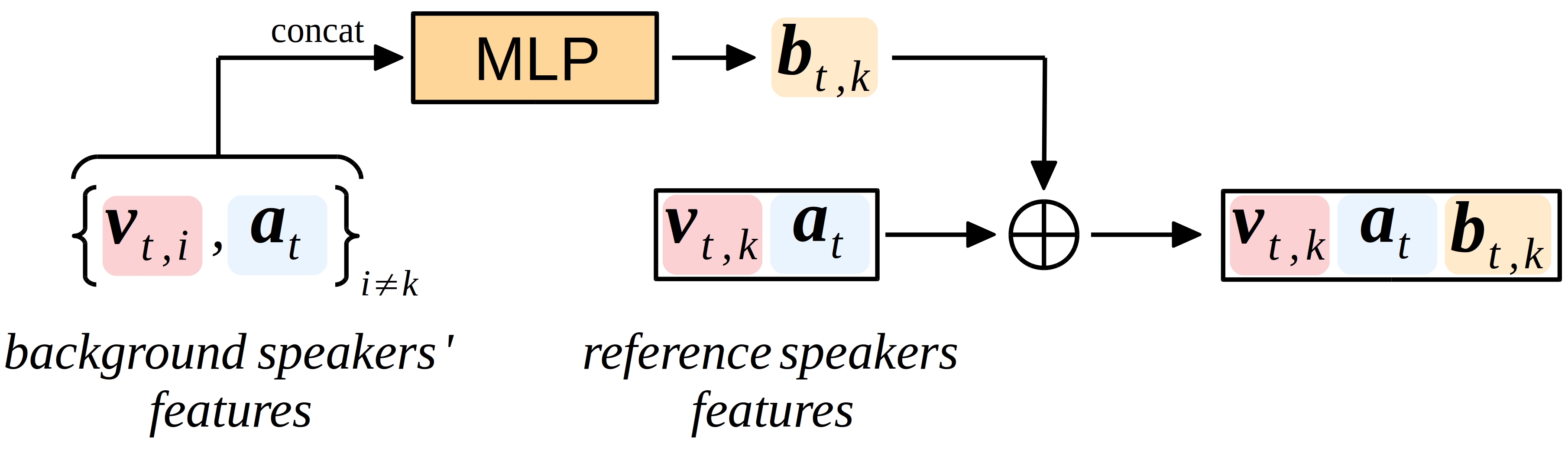}
	\caption{Inter-speaker relation modeling architecture. For reference speaker $k$ at time instant $t$, we extract background features $\mathbf{b_{t,k}}$ by passing the concatenated features of background speakers through one layer MLP. Extracted features are then concatenated to reference speakers video features and audio features.}
	\label{fig:isrm}
\end{figure}

Consider a reference speaker $k$ and $m$ background speakers in the scene at time $t$. The output of the audio-visual encoder for the reference speaker is $[\mathbf{v}_{t,k}, \mathbf{a}_t]$. To incorporate information from background speakers,  we propose to extract an additional feature vector $\mathbf{b}_{t,k}$ using a single-layer perceptron, as illustrated in Fig.~\ref{fig:isrm}. The input to the MLP are the concatenated audio-visual embeddings from all background speakers at time $t$. Note that the number $m$ is fixed from the system's perspective: if there are less than $m$ background speakers at time $t$, the encoder features are populated with zero vectors. If there are more than $m$ speakers, only $m$ are randomly selected. In this manner, the input dimension of the MLP is fixed. The final feature vector $[\mathbf{v}_{t,k}, \mathbf{a}_t, \mathbf{b}_{t,k}]$ is fed to the temporal model. An experimental study of the proposed ISRM, and comparison to the approach in~\cite{alcazar2020active} is provided in Section~\ref{sec:exp_isrm}.

\subsection{Temporal Modeling}
\label{sec:temp_mod}

Speaking is a coherent action in time: if a person is speaking at previous or future time instants, it is likely that the person is speaking at the current time instant. This is also valid for \textit{remaining silent} action. Therefore, temporal modeling is crucial for accurate active speaker detection. 

We experimented with several RNN-based temporal modeling architectures: Long Short-Term Memory (LSTM) \cite{hochreiter1997long}, Gated Recurrent Unit (GRU) \cite{cho2014properties}, Simple Recurrent Unit (SRU) \cite{lei2018sru} and their bidirectional versions. For the uni-directional methods, the reference frame is at the end of the input, while for the bidirectional methods it is at center of the input. The hidden state vector of the recurrent block at the reference frame is fed to a fully connected layer to produce a binary output $\mathbf{z}_t[k] \in \lbrace 0, 1 \rbrace$  (i.e. active speaker or not). In case speakers' features are not available for the selected time window, similar to \cite{alcazar2020active} we apply same padding to the beginning or to the end. Out of all methods, Bidirectional-GRU performs best and becomes our final choice in temporal modeling stage.


\begin{table}[t!]
    \centering
    \begin{tabular}{lcc}
        \specialrule{.1em}{.2em}{.2em}
        \textbf{Audio Backbone} & \textbf{Video Backbone} & \textbf{mAP}  \\ 
        \specialrule{.1em}{.2em}{.2em}
        2D-ResNet-18    & 2D-ResNet-18  & 79.0   \\
        2D-ResNet-18    & 3D-ResNet-18  & 83.9   \\
        SincDSNet       & 2D-ResNet-18  & 80.8     \\
        SincDSNet       & 3D-ResNet-18  & \textbf{86.1}   \\
        \specialrule{.1em}{.2em}{.2em}
    \end{tabular}
    \caption{Performance comparison of different audio and video backbones. Input length of 8-frames is used for all evaluations.}
	\label{tab:comp_backbones}
\end{table}

\begin{table}[t!]
    \centering
    \begin{tabular}{lcc}
        \specialrule{.1em}{.2em}{.2em}
        \textbf{Audio Backbone} & \textbf{Params} & \textbf{MFLOP}  \\ 
        \specialrule{.1em}{.2em}{.2em}
        SincDSNet              & 0.15M   & 13.8     \\
        2D-ResNet-18           & 11.2M   & 19.2     \\
        \specialrule{.1em}{.2em}{.2em}
    \end{tabular}
    \caption{Complexity comparison of different audio backbones.}
	\label{tab:comp_audio_cnn}
\end{table}

\subsection{Training Details}
\label{sec:training}

\paragraph{Training Audio-Visual Encoding Backbones.}
We train our audio-visual encoder using ADAM optimizer \cite{kingma2014adam} for 70 epochs. Batch size is selected as highest possible number that fits to a single NVIDIA Titan XP GPU for different backbones. However, gradients are accumulated reaching to effective batch size of 192 before doing backward propagation. The learning rate is initialized with $3\times 10^{-4}$ and dropped by a factor of 10 at every 30 epochs. For video input, we apply random cropping, random horizontal flipping and color transformations as data augmentation at the training time. Finally, video input is reshaped to the resolution of 160$\times$160. The audio signals are sampled at 16 kHz. 3D-CNNs are pretrained on Kinetics \cite{carreira2017quo}, 2D-CNNs are pretrained with ImageNet \cite{deng2009imagenet}, and SincDSNet is trained from scratch. Once the training is finished, prediction heads are discarded and the features $\mathbf{v}_{t,k} \in \mathbb{R}^{512}$ and $\mathbf{a}_{t} \in \mathbb{R}^{160}$ are used to train the ISRM and the temporal model. 

\paragraph{Training ISRM and Temporal Modeling.} We used ADAM optimizer with cross-entropy loss to train the ISRM and the temporal model. We train for 10 epochs, with batch size of 256. The learning rate is initialized with $3\times 10^{-6}$ and dropped by 10 at 5th epoch. The MLP in the ISRM extracts the feature $\mathbf{b}_{t,k} \in \mathbb{R}^{128}$ independent from the number of background speakers. For the temporal model, we used two recurrent layers with hidden state dimension of 128, which experimentally proved to be optimal for our system. 

Our final architecture ASDNet is implemented in PyTorch and all experiments are performed using a single NVIDIA Titan Xp GPU.

\section{Experiments}
\label{sec:exp}


\paragraph{Dataset.} The AVA-ActiveSpeaker dataset~\cite{roth2020ava} is the first audio-visual active speaker dataset collected in the wild. It contains 262 15-minute videos from Hollywood movies, recorded at 25-30 fps, 120 of which are used for training, 33 for validation, and 109 for testing. The videos consist of 3.65 million human-labeled frames, where face crops belonging to the same speaker are aggregated to create face tracks, and each face crop is annotated with \textit{speaking} or \textit{not-speaking} label. This results in 38.5 hours of face tracks with the corresponding audio signal. The number of speakers in the videos is time-varying, and a significant portion of face crops has resolution less than 100 pixels, making the dataset considerably challenging. 

\vspace{-0.2cm}
\paragraph{Evaluation Metric.} We use the official ActivityNet evaluation tool that computes mean average precision (mAP). Unless stated otherwise, we use AVA-ActiveSpeaker validation set for our evaluations.

\subsection{Audio-Visual Encoder Evaluation}
\label{sec:exp-ave}

In this section, we investigate the advantage of the proposed audio and video backbones, compared to backbones used in state-of-the-art active speaker detection systems. The encoder architecture is of utmost importance: the overall performance of the \ac{AV-ASD} pipeline can only be as good as the extracted features. For these experiments, ISRM and temporal modeling are not used.


\begin{table}[t!]
    \centering
    \begin{tabular}{clccc}
        \specialrule{.1em}{.2em}{.2em}
        & \textbf{Video Backbone} & \textbf{Params} & \textbf{GFLOP}    & \textbf{mAP}   \\ 
        \specialrule{.1em}{.2em}{.2em}
        \multicolumn{1}{c|}{\multirow{2}{*}{\rotatebox[origin=c]{90}{32-f}}} & 3D-ResNeXt-101         & 48.6M   & 13.2  & \textbf{88.9}   \\
        \multicolumn{1}{c|}{} & 3D-ResNet-18  & 33.2M   & 10.3  & 87.4   \\
        \specialrule{.1em}{.1em}{.1em}
        \multicolumn{1}{c|}{\multirow{2}{*}{\rotatebox[origin=c]{90}{16-f}}} & 3D-ResNeXt-101         & 48.6M   & 14.1  & \textbf{88.9}   \\
        \multicolumn{1}{c|}{} & 3D-ResNet-18  & 33.2M   & 11.2  & 87.5   \\
        \specialrule{.1em}{.1em}{.1em}
        \multicolumn{1}{c|}{\multirow{7}{*}{\rotatebox[origin=c]{90}{8-f}}} & 3D-ResNeXt-101         & 48.6M   & 13.2  & 86.7   \\
        \multicolumn{1}{c|}{} & 3D-ResNet-18           & 33.2M   & 10.3  & 86.1   \\
        \multicolumn{1}{c|}{} & 2D-ResNet-18           & 11.2M   & 0.9   & 80.8   \\
        \multicolumn{1}{c|}{} & 3D-MobileNetV1 2.0x    & 13.9M   & 0.6   & 81.6   \\
        \multicolumn{1}{c|}{} & 3D-MobileNetV2 1.0x    & 2.1M    & 0.7   & 85.1   \\
        \multicolumn{1}{c|}{} & 3D-ShuffleNetV1 2.0x   & 4.6M    & 0.7   & 85.0   \\
        \multicolumn{1}{c|}{} & 3D-ShuffleNetV2 2.0x   & 3.9M    & 0.6   & 84.2   \\
        \specialrule{.1em}{.2em}{.2em}
    \end{tabular}
    \caption{Comparison of video backbones for different clip lengths. SincDSNet is used at the audio backbone, and face crop resolution is $160 \times 160$.}
	\label{tab:comp_video_cnn}
\end{table}

\vspace{-0.2cm}
\paragraph{\emph{Which encoder architectures should be used?}}
Following recent works \cite{leon2021maas, alcazar2020active, roth2020ava, chung2019naver, zhangmulti}, we take \mbox{2D-ResNet-18} architecture as the audio and video backbones of a baseline encoder. Inputs to the video backbone are stacked face crops, and inputs to the audio backbone are MFCCs, corresponding to a length of eight frames.
This baseline achieves 79.0 mAP as shown in Table~\ref{tab:comp_backbones}. 

To demonstrate the benefit of applying 3D convolution kernels, we keep the baseline audio backbone and replace 2D-ResNet-18 by 3D-ResNet-18. \emph{This change alone brings improvement of 4.9 mAP over the baseline.} The improvement is achieved solely due to the ability of the 3D convolution kernels to capture motion patterns in the video data.


Similarly, to evaluate the benefit of SincDSNet as the proposed audio backbone, we keep the baseline video backbone and replace the \mbox{`MFCC $+$ 2D-ResNet-18'} audio backbone by SincDSNet. \emph{This change brings improvement of 1.8 mAP over the baseline}, thanks to the partially learnable feature extraction by SincDSNet, operating on the raw audio data.
Importantly, SincDSNet has 75 times less parameters than 2D-ResNet-18 and requires less floating point operations (FLOPs), as shown in Table~\ref{tab:comp_audio_cnn}.

Finally, our audio-visual encoder that uses both 3D-ResNet-18 and SincDSNet as backbones, achieves 7.1 mAP improvement over the baseline.


\begin{table}[t!]
    \centering
    \begin{tabular}{r|cccccc}
        \textbf{\# Speakers}  & \phantom{a} 0  & 1 & 2 & 3 & 4 & 5 \phantom{a}   \\ 
        \specialrule{.1em}{.2em}{.2em}
        \textbf{mAP} & 92.6  & 93.1 & \textbf{93.4} &  \textbf{93.4} & \textbf{93.4} & \textbf{93.3} \\ 
    \end{tabular}
    \vspace{.1cm}
    \caption{Performance of inter-speaker relation modelling for different number of background speakers.}
	\label{tab:background_speakers}
\end{table}

\begin{table}[t!]
    \centering
    \begin{tabular}{lcc}
        \specialrule{.1em}{.2em}{.2em}
        \textbf{Method}  &\textbf{Temporal Model}  & \textbf{mAP} \\ 
        \specialrule{.1em}{.2em}{.2em}
        NonLocal \cite{alcazar2020active}        &              & 87.2 \\
        NonLocal \cite{alcazar2020active}        & \checkmark   & 92.8 \\
        ISRM (ours)      &              & 89.0 \\
        ISRM (ours)      & \checkmark   & \textbf{93.4} \\
        \specialrule{.1em}{.2em}{.2em}
    \end{tabular}
    \vspace{.1cm}
    \caption{Comparison of inter-speakers relation modeling methods.}
	\label{tab:inter-speakers_modeling}
\end{table}

\vspace{-0.3cm}
\paragraph{\emph{Can we use resource-efficient video encoders?}} One can attribute the performance boost achieved by 3D-ResNet-18 backbone to its increased number of parameters and FLOPs. Therefore, we have used several resource efficient 3D CNNs \cite{kopuklu2019resource} as video backbone. We report their performance at the bottom of Table \ref{tab:comp_video_cnn}. Notably, all 3D CNN architectures achieve better performance than 2D-ResNet-18. For isntance, although 3D-MobileNetV2 1.0x contains much smaller number of parameters (approx.~7x less) and less FLOPs compared to 2D-ResNet-18, it achieves around 4 mAP better performance.

We have also experimented with deeper and computationally more expensive 3D-ResNeXt-101 architecture to check how much performance can be increased. 3D-ResNeXt-101 shows 0.6 mAP improvement over 3D-ResNet-18 when 8-frames input is used.

\vspace{-0.3cm}
\paragraph{\emph{How does clip length affect performance?}} Although we used 8-frames clips to train our audio-visual backbones, longer clips would provide larger temporal context. In Table \ref{tab:comp_video_cnn}, we compare clip lengths of 8-frames, 16-frames and 32-frames for the best performing 3D-ResNeXt-101 and 3D-ResNet-18 video backbones. To maintain similar complexity, we removed the initial temporal downsampling for 8-frames input, and inserted an additional temporal downsampling to the initial convolution layer for 32-frames input. Applying 16-frames clip length brings a performance improvement of 1.4 mAP and 2.2 mAP over 8-frames clip length for 3D-ResNet-18 and 3D-ResNeXt-101, respectively. Using 32-frames clip length does not show same performance improvement over using 16-frames. We suspect that inserting additional temporal downsampling hinders backbones ability to capture motion patterns. 


\begin{table}[t!]
    \centering
    \begin{tabular}{lc}
        \specialrule{.1em}{.2em}{.2em}
        \textbf{Background features}  & \textbf{mAP} \\ 
        \specialrule{.1em}{.2em}{.2em}
        only reference frame              & 93.4 \\
        neighbouring window of 9 frames   & \textbf{93.5} \\
        \specialrule{.1em}{.2em}{.2em}
    \end{tabular}
    \caption{Performance comparison when background speakers' features at different number of frames are leveraged.}
	\label{tab:neighbouring-frames}
\end{table}

\subsection{Inter-Speaker Relation Modeling Evaluation}
\label{sec:exp_isrm}

In this section, we investigate the performance of the proposed ISRM and compare it to an existing approach \cite{alcazar2020active} for context modelling. These experiments include the full ASDNet pipeline (encoder, ISRM, and a temporal model), where the temporal model, if present, is a Bidirectional-GRU with sequence length of 64.

\vspace{-0.3cm}
\paragraph{\emph{How many background speakers to use for ISRM?}} We experimented with different number of background speakers for ISRM, and the results are reported in Table \ref{tab:background_speakers}. In general, increasing the number of background speakers features increases the performance.  ISRM increases the performance by 0.8 mAP compared to the case where only reference speaker's features are used with temporal modeling \mbox{(0 background speaker case)}. In the rest of our experiments, we use three background speakers in the ISRM module.

\vspace{-0.3cm}
\paragraph{\emph{How does our ISRM compare to existing approaches?}} In Table~\ref{tab:inter-speakers_modeling}, we provide a comparison of our ISRM approach to the \emph{NonLocal} \cite{wang2018non} approach proposed in \cite{alcazar2020active}. \emph{NonLocal} captures relationships between all the speakers within clip, whereas our ISRM approach captures relationships between speakers only within reference frame. When used alone, after the audio-visual backbones, neither \emph{NonLocal} nor our ISRM approach bring significant performance improvement (\emph{NonLocal} even degrades the performance). However, ISRM contributes additional 0.8 mAP compared to a system that uses only temporal modeling. 

\vspace{-0.3cm}
\paragraph{\emph{Can ISRM benefit from neighbouring frames?}} At ISRM, we do not have to use background speakers' features at only reference frame. Neighbouring frames relative to the reference frame can also provide useful information for ISRM. Therefore we have used background speakers' features at neighbouring window of 9 frames, which shows a modest 0.1 mAP improvement as reported in Table~\ref{tab:neighbouring-frames}. For the rest of the paper, we use 9 neighbouring frames at ISRM.

\begin{table}[t!]
    \centering
    \begin{tabular}{lcc}
        \specialrule{.1em}{.2em}{.2em}
        \textbf{Method} & \textbf{Sequence Length}  & \textbf{mAP}   \\ 
        \specialrule{.1em}{.2em}{.2em}
        Bidirectional-GRU    & 64     & \textbf{93.5} \\
        Bidirectional-LSTM   & 64     & 93.4 \\
        Bidirectional-SRU    & 64     & 93.2 \\
        GRU                  & 32     & 92.8 \\
        LSTM                 & 32     & 92.7 \\
        SRU                  & 32     & 92.7 \\
        \specialrule{.1em}{.2em}{.2em}
    \end{tabular}
    \caption{Performance comparison of temporal modeling methods.}
	\label{tab:temporal_modeling}
\end{table}

\begin{table}[t!]
    \centering
    \begin{tabular}{r|ccccc}
        \textbf{Seq. Length}  & \phantom{a} 8  & 16 & 32 & 64 & 128 \phantom{a}   \\ 
        \specialrule{.1em}{.2em}{.2em}
        \textbf{mAP} & 92.0  & 92.8 & 93.3 &  \textbf{93.5} & \textbf{93.5}\\ 
    \end{tabular}
    \caption{Performance comparison of using different sequence lengths at the training of Bidirectional-GRU.}
	\label{tab:seq_length}
\end{table}

\subsection{Temporal Modeling Evaluation}
\label{sec:exp_temp}

\paragraph{\emph{Which RNN architectures are most suitable?}} Table \ref{tab:temporal_modeling} shows the performance comparison of different RNN blocks used for temporal modeling. All one-directional methods takes 32-frames features as input and last output is used as input to final fc layer (reference frame is placed to the last of input sequence). For bidirectional methods, we have used 64-frames features as input and center output is used as input to final fc layer (reference frame is placed at the center of input sequence). Compared to their bidirectional versions, one-directional methods perform around 0.7 mAP worse. Out of all methods, bidirectional-GRU achieves the best performance.

\vspace{-0.3cm}
\paragraph{\emph{What should be the length of the input sequence?}} We have experimented with different sequence lengths and reported results in Table~\ref{tab:seq_length}. In general, using larger sequence length does not hurt the final performance. However, after sequence length 64, the performance converges to 93.5 mAP.

\begin{figure}[b!]
    \centering
    \subfigure[]{
        \includegraphics[width = 0.47\linewidth]{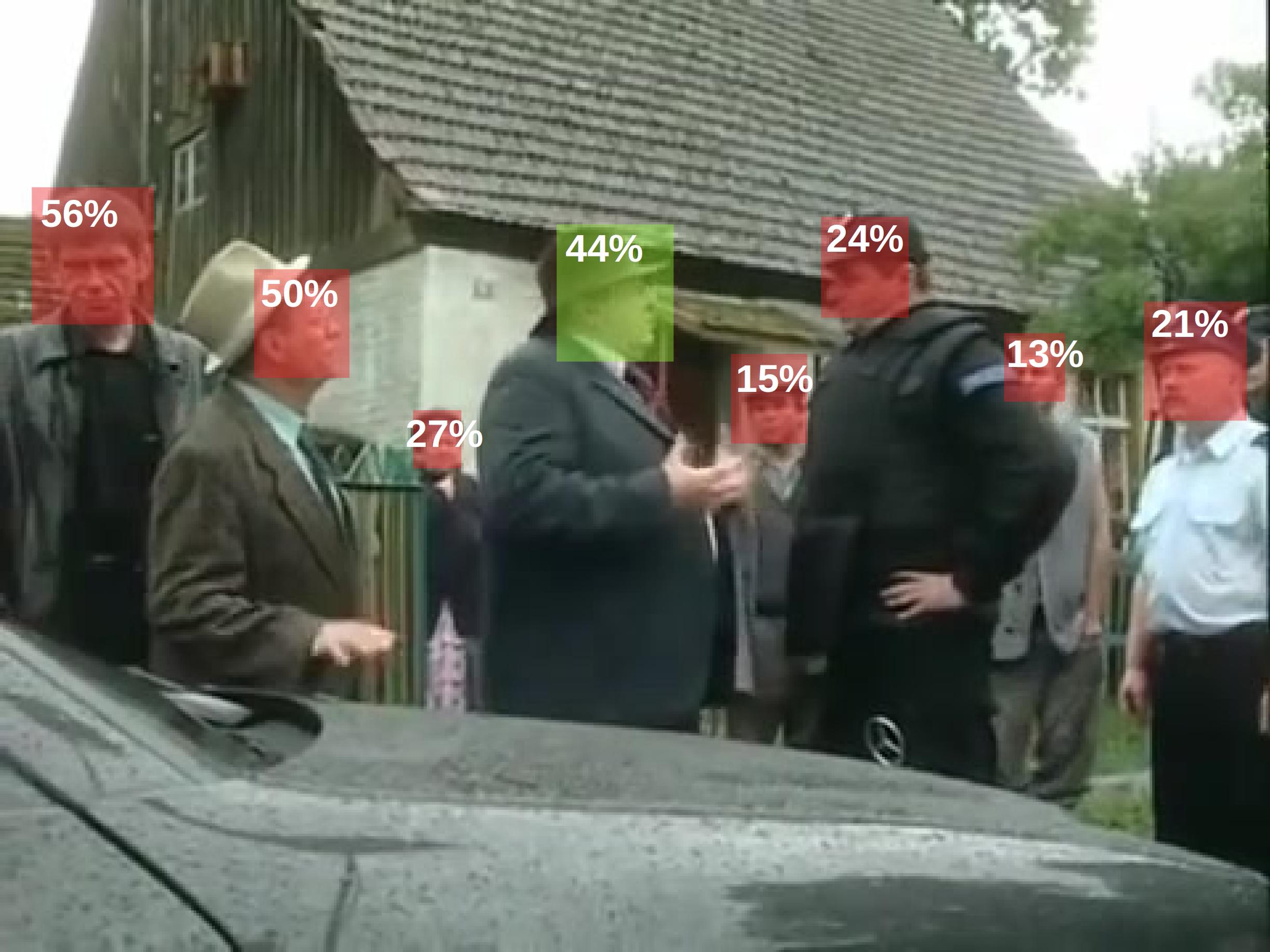}
        } \subfigure[]{
        \includegraphics[width = 0.47\linewidth]{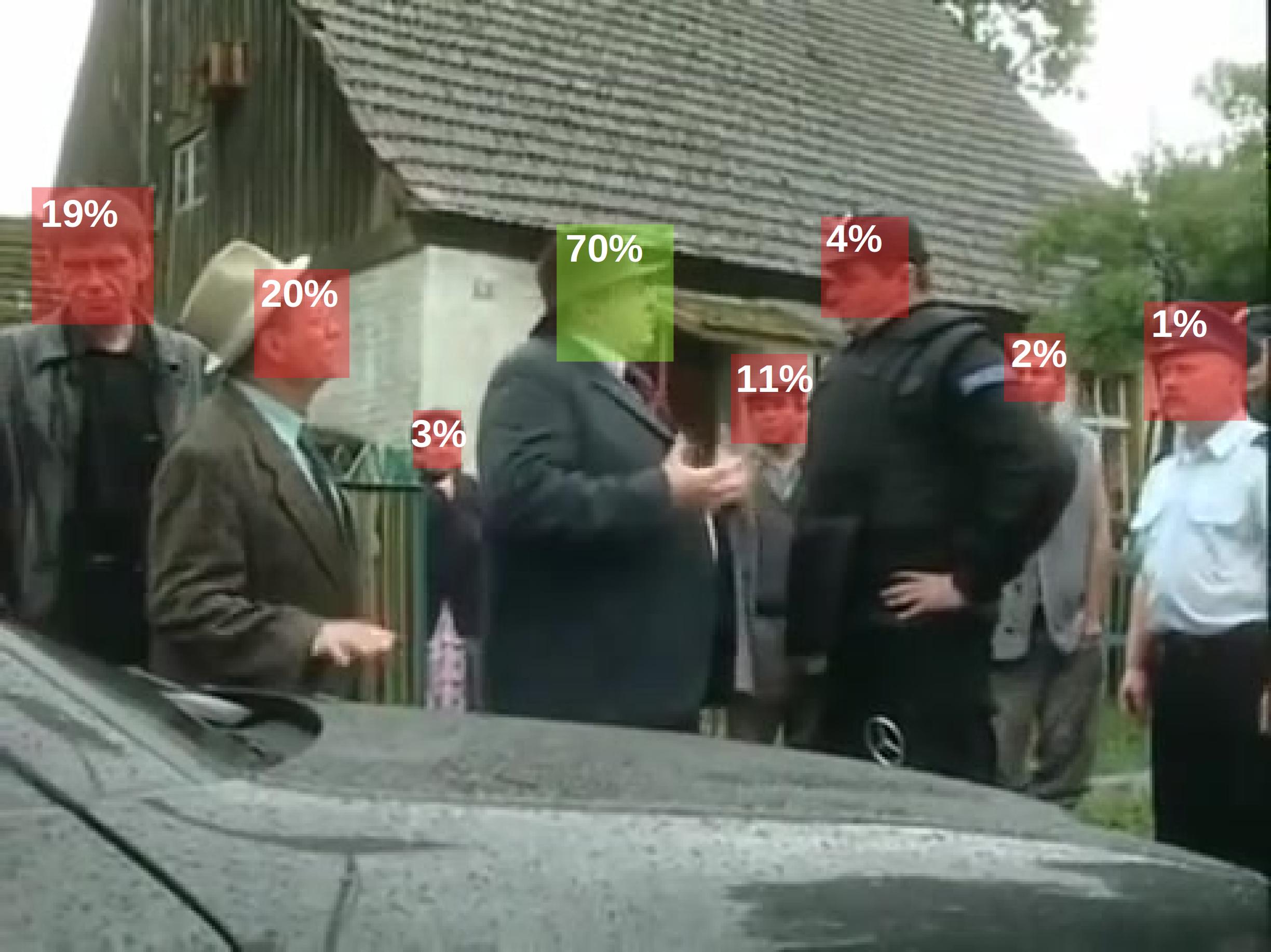}
        }
    \caption{The network predictions for speaking probabilities of each speaker (a) after only audio-visual encoding (b) after temporal modeling and ISRM are also applied. Ground truths of \textit{speaking} and \textit{not-speaking} classes are denoted with green and red rectangles, respectively. }
    \label{fig:visualization}
\end{figure}


\subsection{Component-wise Analysis}

\paragraph{\emph{How does each component contribute to the performance?}} We investigated the contribution of each component to the final performance in Table~\ref{tab:comp-wise}. We highlight several findings: \textbf{(i)} Without ISRM and temporal modeling, suitable backbones alone achieve 88.9 mAP, which is better than any other state-of-the-art approach; \textbf{(ii)} ISRM and temporal modeling improve the performance by 0.7 mAP and 3.7 mAP when they are applied alone, respectively, showing the importance of both stages in the pipeline; \textbf{(iii)} In rows 6 and 7 in Table~\ref{tab:comp-wise}, we investigated the importance of ISRM stage by evaluating the performance without using reference speakers video features. Accordingly, even without looking reference speaker's face, information acquired from background speakers and audio enables our architecture to achieve around 68 mAP. This shows that ISRM is an indispensable part of our pipeline; \textbf{(iv)} When ISRM and temporal modeling are applied together, our architecture achieves the best performance with 93.5 mAP. 

The contribution of temporal modeling and ISRM stages is visually illustrated in Fig.~\ref{fig:visualization}. With only audio-visual encoding, each speaker is analyzed independently and predictions for speaking probabilities are made without contextual and long-term temporal information in Fig.~\ref{fig:visualization}~(a). After applying temporal modeling and ISRM stages, the \mbox{ASDNet} predictions of speaking probabilities for \textit{not-speaking} speakers drop and \textit{speaking} speaker increases considerable as shown in Fig.~\ref{fig:visualization}~(b).

\begin{table}[t!]
    \centering
    \begin{tabular}{cccccc}
        \specialrule{.1em}{.2em}{.2em}
        \textbf{\#} &
        \begin{tabular}[c]{@{}l@{}}\hspace{0.2cm}\textbf{Speaker} \\ \textbf{Video Feat.}\end{tabular} &
        \begin{tabular}[c]{@{}l@{}}\textbf{Audio} \\ \hspace{0.1cm}\textbf{Feat.}\end{tabular} &
        \begin{tabular}[c]{@{}l@{}}\textbf{ISRM} \\ \hspace{0.1cm}\textbf{Feat.}\end{tabular} &
        \begin{tabular}[c]{@{}l@{}}\textbf{Temporal} \\ \hspace{0.0cm}\textbf{Modeling}\end{tabular} &
        \textbf{mAP}   \\ 
        \specialrule{.1em}{.2em}{.2em}
        1 & \cmark   &            &          &            & 78.8 \\
        2 &          & \cmark     &          &            & 49.3 \\
        3 & \cmark   & \cmark     &          &            & 88.9 \\
        4 & \cmark   & \cmark     &          & \cmark     & 92.6 \\
        5 & \cmark   & \cmark     & \cmark   &            & 89.6 \\
        6 &          & \cmark     & \cmark   &            & 64.5 \\
        7 &          & \cmark     & \cmark   & \cmark     & 67.8 \\
        8 & \cmark   & \cmark     & \cmark   & \cmark     & \textbf{93.5}\\
        \specialrule{.1em}{.2em}{.2em}
    \end{tabular}
    \caption{Contribution of each component to the final performance.}
	\label{tab:comp-wise}
\end{table}

\begin{table}[t!]
    \centering
    \begin{tabular}{ccc}
        \specialrule{.1em}{.2em}{.2em}
        \begin{tabular}[c]{@{}l@{}}\hspace{0.3cm}\textbf{Encoder} \\ \textbf{Clip Length}\end{tabular} & \begin{tabular}[c]{@{}l@{}}\hspace{0.7cm}\textbf{ISRM and} \\ \textbf{Temporal Modeling}\end{tabular} & \textbf{mAP}  \\
        \specialrule{.1em}{.2em}{.2em}
        8-frames\phantom{1}   & \xmark    & 86.7      \\
        16-frames             & \xmark    & 88.9      \\
        \specialrule{.1em}{.1em}{.1em}
        8-frames\phantom{1}   & \cmark    & 93.4      \\
        16-frames             & \cmark    & 93.5      \\
        \specialrule{.1em}{.2em}{.2em}
    \end{tabular}
    \caption{Effect of encoder clip length on the final performance. SincDSNet and 3D-ResNeXt-101 are used for audio and video backbones, respectively.}
	\label{tab:enc_clip_length}
\end{table}

\vspace{-0.3cm}
\paragraph{\emph{How does the clip length affect performance?}} Increased encoder clip length (16-frames instead of 8-frames using 3D-ResNeXt-101 video backbone) improves the performance by 2.2 mAP if ISRM and temporal modeling are not applied. However, in the complete pipeline this improvement reflects to a marginal 0.1 mAP improvement in the final performance, which is shown in Table~\ref{tab:enc_clip_length}. This shows that increased encoder clip length shifts the improvement that could have been provided by temporal modeling to the encoder. This might not be desirable if complexity is important at the design of the architecture since doubling encoder clip length means doubling the complexity.   

\vspace{-0.3cm}
\paragraph{\emph{Can ISRM be placed after temporal modeling?}} If necessary, the order of ISRM and temporal modeling can be changed, which results in only a 0.1 mAP performance degradation. 

\vspace{-0.3cm}
\paragraph{\emph{Can we make the full pipeline causal?}} The complete pipeline can be made causal by placing the reference frame to the last place of the input for encoder and temporal modeling stages; and by not using neighbouring frames background speakers' features at ISRM. So that, no future information is used for the active speaker detection of the current frame. Causal pipeline achieves 90.6 mAP, which is still better than any state-of-the-art approach.

\subsection{Comparison with the State-of-the-art}
\label{sec:sota}

\begin{table}[t!]
    \centering
    \begin{tabular}{clr}
        \specialrule{.1em}{.2em}{.2em}
        & \textbf{Method} \phantom{i} & \hspace{1cm} \textbf{mAP} \\ 
        \specialrule{.1em}{.2em}{.2em}
        \multicolumn{1}{c|}{\multirow{7}{*}{\rotatebox[origin=c]{90}{\textit{validation set}}}} & ASDNet (ours)       & \textbf{93.5} \\
        \multicolumn{1}{c|}{} & Causal ASDNet (ours)            & 90.6 \\
        \multicolumn{1}{c|}{} & MAAS-TAN \cite{leon2021maas}            & 88.8 \\
        \multicolumn{1}{c|}{} & Chung et al. \cite{chung2019naver} & 87.8 \\
        \multicolumn{1}{c|}{} & ASC \cite{alcazar2020active} & 87.1 \\
        \multicolumn{1}{c|}{} & Zhang et al. \cite{zhangmulti} & 84.0 \\
        \multicolumn{1}{c|}{} & Sharma et al. \cite{sharma2020crossmodal} & 82.0 \\
        \multicolumn{1}{c|}{} & Roth et al. \cite{roth2020ava} & 79.2 \\
        \specialrule{.1em}{.1em}{.1em}
        \multicolumn{1}{c|}{\multirow{5}{*}{\rotatebox[origin=c]{90}{\textit{test set}}}} & ASDNet (ours)    & \textbf{91.7} \\
        \multicolumn{1}{c|}{} & Chung et al. \cite{chung2019naver} & 87.8 \\
        \multicolumn{1}{c|}{} & ASC \cite{alcazar2020active} & 86.7 \\
        \multicolumn{1}{c|}{} & Zhang et al. \cite{zhangmulti} & 83.5 \\
        \multicolumn{1}{c|}{} & Roth et al. \cite{roth2020ava} & 82.1 \\
        \specialrule{.1em}{.2em}{.2em}
    \end{tabular}
    \caption{Comparison with state-of-the-art methods on the AVA-ActiveSpeaker dataset. mAP results are calculated with the official evaluation tool as explained in \cite{roth2020ava}.}
	\label{tab:sota}
\end{table}

\paragraph{\emph{How does ASDNet compare to state-of-the-art methods?}} We compare the performance of ASDNet with several state-of-the-art methods in Table~\ref{tab:sota}. For the final ASDNet, we used 16-frames clips at the audio-visual encoding stage, 3 background speakers with 9 neighbouring window at the ISRM stage, and bidirectional-GRU with 64-frames sequence length at the temporal modeling stage. ASDNet outperforms the second best approach by 4.7 mAP on the validation set, and by 3.9 mAP on the test set of AVA-ActiveSpeaker dataset.

\vspace{-0.3cm}
\paragraph{\emph{How does number of faces affect the performance?}} Increased number of faces makes the active speaker detection task more challenging and the performance of ISRM becomes more critical. ASDNet outperforms all other state-of-the-art methods for all different face numbers as shown in Table~\ref{tab:num-faces}. Superiority of ASDNet becomes more significant as number of faces increases.

\vspace{-0.3cm}
\paragraph{\emph{How does face size affect the performance?}} Performance comparison for face size, which is set as small for [0, 64), medium for [64, 128), and large for [128, $\infty$) pixels, is shown in Table~\ref{tab:face-size}. ASDNet outperforms all other state-of-the-art methods for all different face sizes. Superiority of ASDNet becomes more significant for smaller faces.

\begin{table}[t!]
    \centering
    \begin{tabular}{lccc}
        \specialrule{.1em}{.2em}{.2em}
        \multicolumn{1}{c}{\multirow{2}{*}{\textbf{Method}}} & \multicolumn{3}{c}{\textbf{Number of Faces}}  \\ \cline{2-4} \addlinespace
        \multicolumn{1}{c}{} & \multicolumn{1}{c}{\phantom{aa}\textbf{1}\phantom{aa}} & \multicolumn{1}{c}{\phantom{aa}\textbf{2}\phantom{aa}} & \multicolumn{1}{c}{\phantom{aa}\textbf{3}\phantom{aa}} \\ 
        \specialrule{.1em}{.2em}{.2em}
        ASDNet (Ours)                    & \textbf{95.7}        & \textbf{92.4}          & \textbf{83.7} \\
        MAAS \cite{leon2021maas}         & 93.3        & 85.8          & 68.2 \\
        ASC \cite{alcazar2020active}     & 91.8        & 83.8          & 67.6 \\
        Baseline \cite{roth2020ava}      & 87.9        & 71.6          & 54.4 \\
        \specialrule{.1em}{.2em}{.2em}
    \end{tabular}
    \caption{Performance comparison by number of visible faces on each frame.}
	\label{tab:num-faces}
\end{table}

\begin{table}[t!]
    \centering
    \begin{tabular}{lccc}
        \specialrule{.1em}{.2em}{.2em}
        \multicolumn{1}{c}{\multirow{2}{*}{\textbf{Method}}} & \multicolumn{3}{c}{\textbf{Face Size}}  \\ \cline{2-4} \addlinespace
        \multicolumn{1}{c}{} & \multicolumn{1}{c}{\phantom{aa}\textbf{S}\phantom{aa}} & \multicolumn{1}{c}{\phantom{aa}\textbf{M}\phantom{aa}} & \multicolumn{1}{c}{\phantom{aa}\textbf{L}\phantom{aa}} \\ 
        \specialrule{.1em}{.2em}{.2em}
        ASDNet (Ours)                    & \textbf{74.3}        & \textbf{89.8}          & \textbf{96.3} \\
        MAAS \cite{leon2021maas}         & 55.2        & 79.4          & 93.0 \\
        ASC \cite{alcazar2020active}     & 56.2        & 79.0          & 92.2 \\
        Baseline \cite{roth2020ava}      & 44.9        & 68.3          & 86.4 \\
        \specialrule{.1em}{.2em}{.2em}
    \end{tabular}
    \caption{Performance comparison by face size.}
	\label{tab:face-size}
\end{table}

\section{Conclusion}
\label{sec:con}

In this paper, we scrutinized the task of Audio-Visual Active Speaker Detection and proposed a three-stage architecture, called ASDNet. With the proposed audio-visual encoder and the inter-speaker relation modelling mechanism, ASDNet outperforms the previous state-of-the-art with significant 4.7 mAP and 3.9 mAP on the validation and test set of AVA-ActiveSpeaker dataset, respectively. To make the final design and hyperparameter choices for ASDNet, we followed insights from carefully designed experiments each targeted a specific aspect of the system. Each of these experiments was discussed in the paper. We believe that these insights can be useful for other complex audio-visual tasks as well that require context and temporal modeling.


{\small
\bibliographystyle{ieee_fullname}
\bibliography{egbib}

\begin{thebibliography}{10}\itemsep=-1pt

\bibitem{alcazar2020active}
Juan~Le{\'o}n Alc{\'a}zar, Fabian Caba, Long Mai, Federico Perazzi, Joon-Young
  Lee, Pablo Arbel{\'a}ez, and Bernard Ghanem.
\newblock Active speakers in context.
\newblock In {\em Proceedings of the IEEE/CVF Conference on Computer Vision and
  Pattern Recognition}, pages 12465--12474, 2020.

\bibitem{Ariav2019}
I. {Ariav} and I. {Cohen}.
\newblock An end-to-end multimodal voice activity detection using wavenet
  encoder and residual networks.
\newblock {\em IEEE Journal of Selected Topics in Signal Processing},
  13(2):265--274, 2019.

\bibitem{carreira2017quo}
Joao Carreira and Andrew Zisserman.
\newblock Quo vadis, action recognition? a new model and the kinetics dataset.
\newblock In {\em proceedings of the IEEE Conference on Computer Vision and
  Pattern Recognition}, pages 6299--6308, 2017.

\bibitem{chakravarty2015s}
Punarjay Chakravarty, Sayeh Mirzaei, Tinne Tuytelaars, and Hugo Van~hamme.
\newblock Who's speaking? audio-supervised classification of active speakers in
  video.
\newblock In {\em Proceedings of the 2015 ACM on International Conference on
  Multimodal Interaction}, pages 87--90, 2015.

\bibitem{cho2014properties}
Kyunghyun Cho, Bart Van~Merri{\"e}nboer, Dzmitry Bahdanau, and Yoshua Bengio.
\newblock On the properties of neural machine translation: Encoder-decoder
  approaches.
\newblock {\em arXiv preprint arXiv:1409.1259}, 2014.

\bibitem{chung2019naver}
Joon~Son Chung.
\newblock Naver at activitynet challenge 2019--task b active speaker detection
  (ava).
\newblock {\em arXiv preprint arXiv:1906.10555}, 2019.

\bibitem{chung2018voxceleb2}
Joon~Son Chung, Arsha Nagrani, and Andrew Zisserman.
\newblock Voxceleb2: Deep speaker recognition.
\newblock {\em Proc. Interspeech 2018}, pages 1086--1090, 2018.

\bibitem{csurka2004visual}
Gabriella Csurka, Christopher Dance, Lixin Fan, Jutta Willamowski, and
  C{\'e}dric Bray.
\newblock Visual categorization with bags of keypoints.
\newblock In {\em Workshop on statistical learning in computer vision, ECCV},
  volume~1, pages 1--2. Prague, 2004.

\bibitem{cutler2000look}
Ross Cutler and Larry Davis.
\newblock Look who's talking: Speaker detection using video and audio
  correlation.
\newblock In {\em 2000 IEEE International Conference on Multimedia and Expo.
  ICME2000. Proceedings. Latest Advances in the Fast Changing World of
  Multimedia (Cat. No. 00TH8532)}, volume~3, pages 1589--1592. IEEE, 2000.

\bibitem{darrell2000audio}
Trevor Darrell, John~W Fisher, and Paul Viola.
\newblock Audio-visual segmentation and “the cocktail party effect”.
\newblock In {\em International Conference on Multimodal Interfaces}, pages
  32--40. Springer, 2000.

\bibitem{deng2009imagenet}
Jia Deng, Wei Dong, Richard Socher, Li-Jia Li, Kai Li, and Li Fei-Fei.
\newblock Imagenet: A large-scale hierarchical image database.
\newblock In {\em 2009 IEEE conference on computer vision and pattern
  recognition}, pages 248--255. Ieee, 2009.

\bibitem{dieleman2014end}
Sander Dieleman and Benjamin Schrauwen.
\newblock End-to-end learning for music audio.
\newblock In {\em 2014 IEEE International Conference on Acoustics, Speech and
  Signal Processing (ICASSP)}, pages 6964--6968. IEEE, 2014.

\bibitem{Ephrat2018}
Ariel Ephrat, Inbar Mosseri, Oran Lang, Tali Dekel, Kevin Wilson, Avinatan
  Hassidim, William~T. Freeman, and Michael Rubinstein.
\newblock Looking to listen at the cocktail party: A speaker-independent
  audio-visual model for speech separation.
\newblock {\em ACM Trans. Graph.}, July 2018.

\bibitem{feichtenhofer2019slowfast}
Christoph Feichtenhofer, Haoqi Fan, Jitendra Malik, and Kaiming He.
\newblock Slowfast networks for video recognition.
\newblock In {\em Proceedings of the IEEE/CVF International Conference on
  Computer Vision}, pages 6202--6211, 2019.

\bibitem{gao2020listen}
Ruohan Gao, Tae-Hyun Oh, Kristen Grauman, and Lorenzo Torresani.
\newblock Listen to look: Action recognition by previewing audio.
\newblock In {\em Proceedings of the IEEE/CVF Conference on Computer Vision and
  Pattern Recognition}, pages 10457--10467, 2020.

\bibitem{gebru2017audio}
Israel~D Gebru, Sileye Ba, Xiaofei Li, and Radu Horaud.
\newblock Audio-visual speaker diarization based on spatiotemporal bayesian
  fusion.
\newblock {\em IEEE transactions on pattern analysis and machine intelligence},
  40(5):1086--1099, 2017.

\bibitem{hara2018can}
Kensho Hara, Hirokatsu Kataoka, and Yutaka Satoh.
\newblock Can spatiotemporal 3d cnns retrace the history of 2d cnns and
  imagenet?
\newblock In {\em Proceedings of the IEEE conference on Computer Vision and
  Pattern Recognition}, pages 6546--6555, 2018.

\bibitem{hershey2016deep}
John~R Hershey, Zhuo Chen, Jonathan Le~Roux, and Shinji Watanabe.
\newblock Deep clustering: Discriminative embeddings for segmentation and
  separation.
\newblock In {\em 2016 IEEE International Conference on Acoustics, Speech and
  Signal Processing (ICASSP)}, pages 31--35. IEEE, 2016.

\bibitem{hochreiter1997long}
Sepp Hochreiter and J{\"u}rgen Schmidhuber.
\newblock Long short-term memory.
\newblock {\em Neural computation}, 9(8):1735--1780, 1997.

\bibitem{ji20123d}
Shuiwang Ji, Wei Xu, Ming Yang, and Kai Yu.
\newblock 3d convolutional neural networks for human action recognition.
\newblock {\em IEEE transactions on pattern analysis and machine intelligence},
  35(1):221--231, 2012.

\bibitem{karpathy2014large}
Andrej Karpathy, George Toderici, Sanketh Shetty, Thomas Leung, Rahul
  Sukthankar, and Li Fei-Fei.
\newblock Large-scale video classification with convolutional neural networks.
\newblock In {\em Proceedings of the IEEE conference on Computer Vision and
  Pattern Recognition}, pages 1725--1732, 2014.

\bibitem{kingma2014adam}
Diederik~P Kingma and Jimmy Ba.
\newblock Adam: A method for stochastic optimization.
\newblock {\em arXiv preprint arXiv:1412.6980}, 2014.

\bibitem{kopuklu2019comparative}
Okan K{\"o}p{\"u}kl{\"u}, Fabian Herzog, and Gerhard Rigoll.
\newblock Comparative analysis of cnn-based spatiotemporal reasoning in videos.
\newblock {\em arXiv preprint arXiv:1909.05165}, 2019.

\bibitem{kopuklu2019resource}
Okan Kopuklu, Neslihan Kose, Ahmet Gunduz, and Gerhard Rigoll.
\newblock Resource efficient 3d convolutional neural networks.
\newblock In {\em Proceedings of the IEEE/CVF International Conference on
  Computer Vision Workshops}, pages 0--0, 2019.

\bibitem{kopuklu2018motion}
Okan Kopuklu, Neslihan Kose, and Gerhard Rigoll.
\newblock Motion fused frames: Data level fusion strategy for hand gesture
  recognition.
\newblock In {\em Proceedings of the IEEE Conference on Computer Vision and
  Pattern Recognition Workshops}, pages 2103--2111, 2018.

\bibitem{krizhevsky2012imagenet}
Alex Krizhevsky, Ilya Sutskever, and Geoffrey~E Hinton.
\newblock Imagenet classification with deep convolutional neural networks.
\newblock {\em Advances in neural information processing systems},
  25:1097--1105, 2012.

\bibitem{kurzinger2020lightweight}
Ludwig K{\"u}rzinger, Nicolas Lindae, Palle Klewitz, and Gerhard Rigoll.
\newblock Lightweight end-to-end speech recognition from raw audio data using
  sinc-convolutions.
\newblock {\em Proc. Interspeech 2020}, pages 1659--1663, 2020.

\bibitem{laptev2005space}
Ivan Laptev.
\newblock On space-time interest points.
\newblock {\em International journal of computer vision}, 64(2-3):107--123,
  2005.

\bibitem{laptev2008learning}
Ivan Laptev, Marcin Marszalek, Cordelia Schmid, and Benjamin Rozenfeld.
\newblock Learning realistic human actions from movies.
\newblock In {\em 2008 IEEE Conference on Computer Vision and Pattern
  Recognition}, pages 1--8. IEEE, 2008.

\bibitem{lee2017raw}
Jongpil Lee, Taejun Kim, Jiyoung Park, and Juhan Nam.
\newblock Raw waveform-based audio classification using sample-level cnn
  architectures.
\newblock {\em arXiv preprint arXiv:1712.00866}, 2017.

\bibitem{lei2018sru}
Tao Lei, Yu Zhang, Sida~I. Wang, Hui Dai, and Yoav Artzi.
\newblock Simple recurrent units for highly parallelizable recurrence.
\newblock In {\em Empirical Methods in Natural Language Processing (EMNLP)},
  2018.

\bibitem{leon2021maas}
Juan Le{\'o}n-Alc{\'a}zar, Fabian~Caba Heilbron, Ali Thabet, and Bernard
  Ghanem.
\newblock Maas: Multi-modal assignation for active speaker detection.
\newblock {\em arXiv preprint arXiv:2101.03682}, 2021.

\bibitem{liu2020multichannel}
Chang-Le Liu, Sze-Wei Fu, You-Jin Li, Jen-Wei Huang, Hsin-Min Wang, and Yu
  Tsao.
\newblock Multichannel speech enhancement by raw waveform-mapping using fully
  convolutional networks.
\newblock {\em IEEE/ACM Transactions on Audio, Speech, and Language
  Processing}, 28:1888--1900, 2020.

\bibitem{materzynska2019jester}
Joanna Materzynska, Guillaume Berger, Ingo Bax, and Roland Memisevic.
\newblock The jester dataset: A large-scale video dataset of human gestures.
\newblock In {\em Proceedings of the IEEE/CVF International Conference on
  Computer Vision Workshops}, pages 0--0, 2019.

\bibitem{miao2017multimodal}
Qiguang Miao, Yunan Li, Wanli Ouyang, Zhenxin Ma, Xin Xu, Weikang Shi, and
  Xiaochun Cao.
\newblock Multimodal gesture recognition based on the resc3d network.
\newblock In {\em Proceedings of the IEEE International Conference on Computer
  Vision Workshops}, pages 3047--3055, 2017.

\bibitem{mittermaier2020small}
Simon Mittermaier, Ludwig K{\"u}rzinger, Bernd Waschneck, and Gerhard Rigoll.
\newblock Small-footprint keyword spotting on raw audio data with
  sinc-convolutions.
\newblock In {\em ICASSP 2020-2020 IEEE International Conference on Acoustics,
  Speech and Signal Processing (ICASSP)}, pages 7454--7458. IEEE, 2020.

\bibitem{monfort2019moments}
Mathew Monfort, Alex Andonian, Bolei Zhou, Kandan Ramakrishnan, Sarah~Adel
  Bargal, Tom Yan, Lisa Brown, Quanfu Fan, Dan Gutfreund, Carl Vondrick, et~al.
\newblock Moments in time dataset: one million videos for event understanding.
\newblock {\em IEEE transactions on pattern analysis and machine intelligence},
  42(2):502--508, 2019.

\bibitem{nagrani2020speech2action}
Arsha Nagrani, Chen Sun, David Ross, Rahul Sukthankar, Cordelia Schmid, and
  Andrew Zisserman.
\newblock Speech2action: Cross-modal supervision for action recognition.
\newblock In {\em Proceedings of the IEEE/CVF Conference on Computer Vision and
  Pattern Recognition}, pages 10317--10326, 2020.

\bibitem{perronnin2010improving}
Florent Perronnin, Jorge S{\'a}nchez, and Thomas Mensink.
\newblock Improving the fisher kernel for large-scale image classification.
\newblock In {\em European conference on computer vision}, pages 143--156.
  Springer, 2010.

\bibitem{qiu2017learning}
Zhaofan Qiu, Ting Yao, and Tao Mei.
\newblock Learning spatio-temporal representation with pseudo-3d residual
  networks.
\newblock In {\em proceedings of the IEEE International Conference on Computer
  Vision}, pages 5533--5541, 2017.

\bibitem{ravanelli2018speaker}
Mirco Ravanelli and Yoshua Bengio.
\newblock Speaker recognition from raw waveform with sincnet.
\newblock In {\em 2018 IEEE Spoken Language Technology Workshop (SLT)}, pages
  1021--1028. IEEE, 2018.

\bibitem{roth2020ava}
Joseph Roth, Sourish Chaudhuri, Ondrej Klejch, Radhika Marvin, Andrew
  Gallagher, Liat Kaver, Sharadh Ramaswamy, Arkadiusz Stopczynski, Cordelia
  Schmid, Zhonghua Xi, et~al.
\newblock Ava active speaker: An audio-visual dataset for active speaker
  detection.
\newblock In {\em ICASSP 2020-2020 IEEE International Conference on Acoustics,
  Speech and Signal Processing (ICASSP)}, pages 4492--4496. IEEE, 2020.

\bibitem{russakovsky2015imagenet}
Olga Russakovsky, Jia Deng, Hao Su, Jonathan Krause, Sanjeev Satheesh, Sean Ma,
  Zhiheng Huang, Andrej Karpathy, Aditya Khosla, Michael Bernstein, et~al.
\newblock Imagenet large scale visual recognition challenge.
\newblock {\em International journal of computer vision}, 115(3):211--252,
  2015.

\bibitem{sharma2020crossmodal}
Rahul Sharma, Krishna Somandepalli, and Shrikanth Narayanan.
\newblock Crossmodal learning for audio-visual speech event localization.
\newblock {\em arXiv preprint arXiv:2003.04358}, 2020.

\bibitem{simonyan2014two}
Karen Simonyan and Andrew Zisserman.
\newblock Two-stream convolutional networks for action recognition in videos.
\newblock In {\em Proceedings of the 27th International Conference on Neural
  Information Processing Systems-Volume 1}, pages 568--576, 2014.

\bibitem{stoller2018waveunet}
Daniel Stoller, Sebastian Ewert, and Simon Dixon.
\newblock Wave-u-net: {A} multi-scale neural network for end-to-end audio
  source separation.
\newblock In {\em Proceedings of the 19th International Society for Music
  Information Retrieval Conference, {ISMIR} 2018, Paris, France, September
  23-27, 2018}, pages 334--340, 2018.

\bibitem{tran2015learning}
Du Tran, Lubomir Bourdev, Rob Fergus, Lorenzo Torresani, and Manohar Paluri.
\newblock Learning spatiotemporal features with 3d convolutional networks.
\newblock In {\em Proceedings of the IEEE international conference on computer
  vision}, pages 4489--4497, 2015.

\bibitem{tran2018closer}
Du Tran, Heng Wang, Lorenzo Torresani, Jamie Ray, Yann LeCun, and Manohar
  Paluri.
\newblock A closer look at spatiotemporal convolutions for action recognition.
\newblock In {\em Proceedings of the IEEE conference on Computer Vision and
  Pattern Recognition}, pages 6450--6459, 2018.

\bibitem{wang2013action}
Heng Wang and Cordelia Schmid.
\newblock Action recognition with improved trajectories.
\newblock In {\em Proceedings of the IEEE international conference on computer
  vision}, pages 3551--3558, 2013.

\bibitem{wang2019voicefilter}
Quan Wang, Hannah Muckenhirn, Kevin Wilson, Prashant Sridhar, Zelin Wu, John~R
  Hershey, Rif~A Saurous, Ron~J Weiss, Ye Jia, and Ignacio~Lopez Moreno.
\newblock Voicefilter: Targeted voice separation by speaker-conditioned
  spectrogram masking.
\newblock {\em Proc. Interspeech 2019}, pages 2728--2732, 2019.

\bibitem{wang2018non}
Xiaolong Wang, Ross Girshick, Abhinav Gupta, and Kaiming He.
\newblock Non-local neural networks.
\newblock In {\em Proceedings of the IEEE conference on computer vision and
  pattern recognition}, pages 7794--7803, 2018.

\bibitem{xu2015empirical}
Bing Xu, Naiyan Wang, Tianqi Chen, and Mu Li.
\newblock Empirical evaluation of rectified activations in convolutional
  network.
\newblock {\em arXiv preprint arXiv:1505.00853}, 2015.

\bibitem{zeghidour2018learning}
Neil Zeghidour, Nicolas Usunier, Iasonas Kokkinos, Thomas Schaiz, Gabriel
  Synnaeve, and Emmanuel Dupoux.
\newblock Learning filterbanks from raw speech for phone recognition.
\newblock In {\em 2018 IEEE international conference on acoustics, speech and
  signal Processing (ICASSP)}, pages 5509--5513. IEEE, 2018.

\bibitem{zhangmulti}
Yuan-Hang Zhang, Jingyun Xiao, Shuang Yang, and Shiguang Shan.
\newblock Multi-task learning for audio-visual active speaker detection.

\end{thebibliography}
}

\end{document}